\documentclass[runningheads]{llncs}
\usepackage[T1]{fontenc}
\usepackage{graphicx}
\begin{document}
\title{Multi-Masked Querying Network for Robust Emotion Recognition from Incomplete Multi-Modal Physiological Signals}
\titlerunning{MMQ-Net for Emotion Recognition from Incomplete Multi-Modal Data}
%
\author{Geng-Xin Xu\inst{1}\orcidID{0000-0001-5710-8643} \and
Xiang Zuo\inst{1,2}\orcidID{0009-0003-7933-3982} \and
Ye Li\inst{1}\thanks{corresponding author.}\orcidID{0000-0002-5351-8546}
}
%
\authorrunning{G. Xu et al.}
%
\institute{Shenzhen Institutes of Advanced Technology, Chinese Academy of Sciences, Shenzhen 518055, China
\and
Southern University of Science and Technology, Shenzhen 518055, China\\
\email{\{gx.xu,x.zuo,ye.li\}@siat.ac.cn}}
\maketitle              
\begin{abstract}
Emotion recognition from physiological data is crucial for mental health assessment, yet it faces two significant challenges: incomplete multi-modal signals and interference from body movements and artifacts. This paper presents a novel Multi-Masked Querying Network (MMQ-Net) to address these issues by integrating multiple querying mechanisms into a unified framework. Specifically, it uses modality queries to reconstruct missing data from incomplete signals, category queries to focus on emotional state features, and interference queries to separate relevant information from noise. Extensive experiment results demonstrate the superior emotion recognition performance of MMQ-Net compared to existing approaches, particularly under high levels of data incompleteness.

\end{abstract}
\section{Introduction}

Mental disorders, such as anxiety disorders and post-traumatic stress disorder, are often accompanied by dysfunctions in emotion processing, which can impair individuals' social abilities and quality of life~\cite{wilmer2021correlates,miller2024quality,lin2025brain}. Neurobiological studies have demonstrated that emotion generation is closely associated with the activity of brain regions such as the limbic system and the prefrontal cortex. Moreover, the interactions between these brain regions and the peripheral physiological system are reflected in physiological indicators such as heart rate, skin conductance response, and respiration~\cite{shu2018review}. Therefore, accurately identifying emotional states is crucial for mental health assessment and intervention.

With advancements in electroencephalography (EEG) and peripheral physiological signal monitoring technologies, emotion recognition methods based on multi-modal physiological signals provide more objective and real-time emotion monitoring tools for clinical applications~\cite{zhu2023dynamic,zhang2024camel,shou2024adversarial,wang2025generative}. In recent years, many researchers have explored emotion recognition using multi-modal approaches~\cite{liu2023emotionkd,tang2024hierarchical,jia2024multi}. Despite the promising results of these studies, there remain two main challenges in this field.

\begin{figure}[t]
\includegraphics[width=\textwidth]{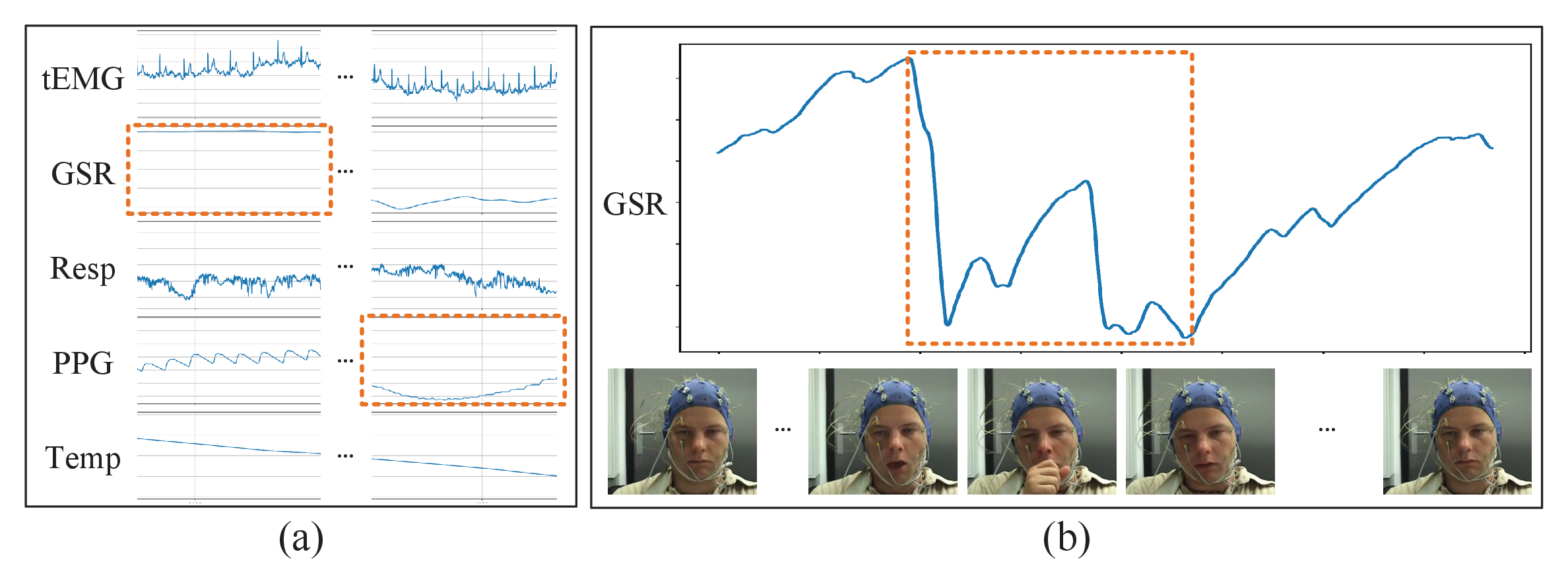}
\caption{Challenges in robust emotion recognition from incomplete multi-modal data. (a) Incomplete signals: Physiological signals with missing segments (dashed boxes), highlighting the incomplete learning problem. (b) Interference: Body movements and artifacts, complicating accurate emotion recognition.}\label{fig1}
\end{figure}

\textbf{Incomplete Multi-Modal Signals.} A key challenge in emotion recognition from physiological signals is the incomplete multi-modal signals. Multi-modal emotion recognition typically involves combining physiological signals from different sources. However, in practical scenarios, these signals are often incomplete due to technical issues, sensor malfunctions, or loss of data during transmission~\cite{liu2024contrastive,lian2023gcnet}. For instance, the galvanic skin response (GSR) or photoplethysmography (PPG) signals may have missing values or even entire segments that are unusable. As shown in Fig.~\ref{fig1}(a), the signals from different modalities can have gaps (indicated by dashed boxes), which result in the incomplete data. This problem significantly hampers the learning process.

\textbf{Interference from Body Movements and Artifacts.} Another major challenge is interference caused by body movements and other external artifacts. When individuals move their body, cough, or experience any other physical disturbance, it can generate noise in the physiological signals, especially in EEG and GSR data. This type of interference can be caused by external factors such as improper sensor placement, muscle contractions, or environmental disturbances~\cite{miao2021automated}. As illustrated in Fig.~\ref{fig1}(b), body movements and artifacts complicate accurate emotion recognition by introducing noise into the signal data, which reduces the signal's reliability. This issue is particularly troublesome when working with real-time or in-the-wild emotion recognition systems, where controlling or eliminating all physical movements is impractical. Consequently, artifacts can lead to incorrect or ambiguous conclusions about the emotional state, further complicating the task of robust emotion recognition.

To address these challenges, we propose a Multi-Masked Querying Network (MMQ-Net) for robust emotion recognition from incomplete multi-modal physiological signals. The core idea of MMQ-Net is to utilize multiple queries within a single framework. Specifically, to handle incomplete multi-modal signals, MMQ-Net uses masked modality queries to reconstruct missing data from the available modalities. To address the interference problem, it incorporates masked category query and interference query to separate emotional state features from irrelevant noise. By combining these three masked querying mechanisms in a unified framework, MMQ-Net is able to robustly handle missing data and interference, thereby improving the accuracy and reliability of emotion recognition in challenging settings.

Overall, the main contributions of this work can be summarized as follows. (1) A novel method named MMQ-Net, is proposed for robust emotion recognition from multi-modal physiological signals that are affected by missing data and interference noise. (2) A multi-masked querying transformer is designed to simultaneously reconstruct incomplete multi-modal features and reduce interference from emotion-irrelevant information, thereby enhancing the robustness of emotion recognition. 

\section{Method}

\subsection{Overview}
Let $\{(\mathbf{x}_i, \mathbf{a}_i, \mathbf{y}_i)\}_{i=1}^n$ denote the multi-modal data with a sample size of $n$, where $\mathbf{x}_i = [\mathbf{x}_i^{(1)}, \ldots, \mathbf{x}_i^{(M)}]$ represents the $M$ modalities of physiological signals, and $\mathbf{x}_i^{(m)}$ is the data from the $m$-th modality. The vector $\mathbf{a}_i \in \{0, 1\}^M$ indicates whether the $M$ modalities are present for sample $i$ (1 if available, 0 if missing). The vector $\mathbf{y}_i$ is a one-hot categorical variable that indicates the specific emotional state. The task of this work is to develop a robust model to predict the emotional state $\mathbf{y}$, considering that the multi-modal physiological signals $\mathbf{x}$ may contain missing data and interference noise.

\begin{figure}[t]
\includegraphics[width=\textwidth]{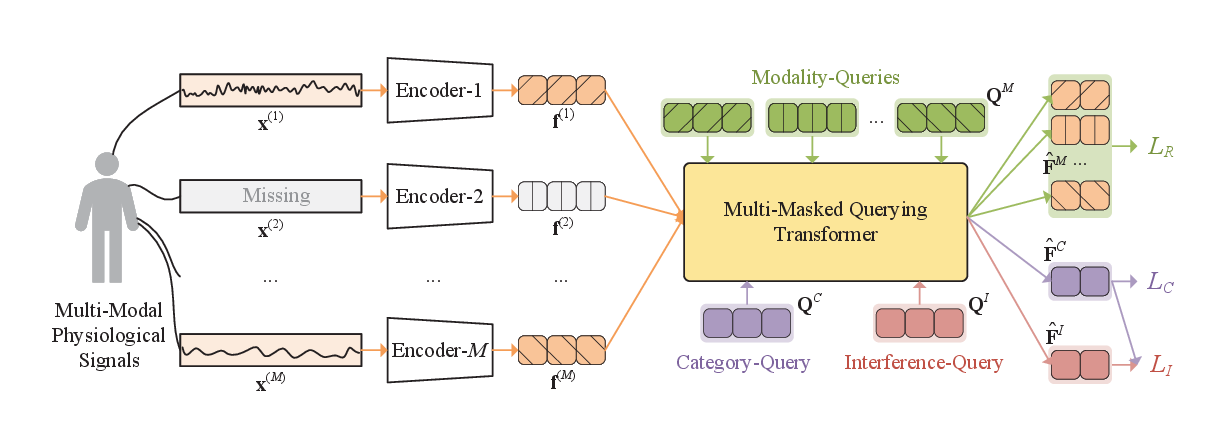}
\caption{Flowchart of the MMQ-Net. Multi-model physiological signals are processed via encoders to extract features, which are then input into a Multi-Masked Querying Transformer to handle incomplete data and reduce interference.} \label{fig2}
\end{figure}

An overview of the proposed MMQ-Net is shown in Fig.~\ref{fig2}. The input includes various physiological signals, such as EEG, GSR, PPG, etc. These signals are processed through respective encoders to extract multi-modal features. The encoded feature vectors are then fed into a Multi-Masked Quering Transformer for the learning of incomplete multi-modal information and reduction of interference. The output is used to compute three loss functions: the multi-modal learning loss function $\mathcal{L}_R$, the discriminative learning loss function $\mathcal{L}_C$, and the interference reduction loss function $\mathcal{L}_I$.

\subsection{Feature Extraction}
In the preprocessing stage of physiological signals, we first perform data cleaning and filtering based on Steve Luck's procedures~\cite{luck2014introduction} to exclude noise components. A notch filter is applied to remove the 50 Hz power line interference, and a 4-45 Hz bandpass filter is used to further clean the data, reducing the impact of measurement inaccuracies and environmental noise. Independent component analysis is then used to remove noise from signals such as electrooculogram (EOG), electrocardiogram (ECG), and electromyogram (EMG). Finally, the original data is downsampled to 128 Hz to reduce the data volume and accelerate computation. The above operations are used for EEG and other physiological signals.

For feature extraction, two methods are primarily used: differential entropy (DE) and power spectral density (PSD). These methods enhance feature representation across five frequency bands: $\theta$ (4-7 Hz), $\alpha$ (8-10 Hz), slow $\alpha$ (8-13 Hz), $\beta$ (14-29 Hz), and $\gamma$ (30-45 Hz). DE extraction assumes a Gaussian distribution, with the standard deviation computed every 2 seconds as the DE value. PSD extraction uses the Welch method, applying the Hanning window function to segment the signal and perform a fast Fourier transform to calculate the power spectral density for each frequency band. Finally, the DE and PSD features are fused using the multi-head attention mechanism~\cite{zhao2024feature} to generate a new feature vector for subsequent analysis.

For simplicity, let $f_m$ represent the feature extraction function for modality $m$, yielding the feature $\mathbf{f}^{(m)} = f_m(\mathbf{x}^{(m)})$. After feature extraction across all modalities, we obtain $\mathbf{F}^M = [\mathbf{f}^{(1)}, \ldots, \mathbf{f}^{(M)}]$.

\subsection{Multi-Masked Querying Transformer}
The Multi-Masked Querying Transformer is designed to handle incomplete multi-modal physiological representations and reduce emotion-irrelevant information. Specifically, this module uses multiple masked queries: modality queries ($\mathbf{Q}^M$), category queries ($\mathbf{Q}^C$), and interference queries ($\mathbf{Q}^I$). Masked modality queries are used to learn missing modalities from available ones, while masked category and interference queries are used to learn features related to emotional states and irrelevant features, respectively.

To learn multi-modal features from incomplete physiological representations, modality queries $\mathbf{Q}^M$ are used as learnable parameters to replace missing data in $\mathbf{F}^M$, encouraging modality completion. To perform attention computation within a unified framework, category queries $\mathbf{Q}^C$ and interference queries $\mathbf{Q}^I$ are concatenated as additional tokens with the features, yielding the following representation:
\begin{equation}\label{equ:F'}
\mathbf{F}' = [\mathbf{F}^M \odot \mathbf{a} + \mathbf{Q}^M \odot (1 - \mathbf{a}), \mathbf{Q}^C, \mathbf{Q}^I],
\end{equation}
where $\odot$ denotes the Hadamard product, and the concatenation operation is implemented at the modality level.

To prevent those missing modalities from affecting attention mechanism, a mask matrix is introduced, ensuring that the queries only learn representations from the available modalities. Let $\mathbf{D}$ be the identity matrix, and $\mathbf{1}$ the vector of ones. The attention mask matrix $\mathbf{M}$ is derived from the modality indicator vector $\mathbf{a}$ as follows:
\begin{equation}\label{equ:M}
\mathbf{M} = \mathbf{D} + \mathbf{1} [\mathbf{a}^{\mathsf{T}}, 1, 1],
\end{equation}
where the last two elements of $1$ correspond to the category query $\mathbf{Q}^C$ and interference query $\mathbf{Q}^I$. Based on the multiple queries and the modality mask matrix, the attention mechanism in the multi-masked querying transformer is computed as:
\begin{equation}\label{equ:Atten}
\mathbf{Z} = \mathrm{softmax}\left(\frac{q(\mathbf{F}') [k(\mathbf{F}')]^T}{\sqrt{d}}\mathbf{M}\right) v(\mathbf{F}').
\end{equation}
This process uses the multi-head attention mechanism~\cite{vaswani2017attention}. For convenience, the output $\mathbf{Z}$ is decomposed into three parts, namely single-modality features $\hat{\mathbf{F}}^M$, emotional state features $\hat{\mathbf{F}}^C$, and interference features $\hat{\mathbf{F}}^I$:
\begin{equation}\label{equ:Z}
\mathbf{Z} := [\hat{\mathbf{F}}^M, \hat{\mathbf{F}}^C, \hat{\mathbf{F}}^I].
\end{equation}

\subsection{Objective Function}
The objective function of MMQ-Net consists of three terms: multi-modal reconstruction loss $\mathcal{L}_R$, discriminative learning loss $\mathcal{L}_C$, and interference reduction loss $\mathcal{L}_I$. For the first term, we perform multi-modal feature reconstruction at the feature level:
\begin{equation}\label{equ:L_R}
\mathcal{L}_R = \frac{1}{\sum_i \mathbf{a}_i^T \mathbf{1}} \sum_i \| \mathbf{a}_i \odot \hat{\mathbf{F}}_i^M - \mathbf{a}_i \odot \mathbf{F}_i^M \|^2.
\end{equation}
For the second term, emotional state features $\hat{\mathbf{F}}^C$ are passed through a multi-layer perceptron (MLP) to obtain the final predicted result, which is then used to compute the cross-entropy loss with the ground truth label $\mathbf{y}$:
\begin{equation}\label{equ:L_C}
\mathcal{L}_C = \frac{1}{n} \sum\nolimits_i \mathrm{CE}[\mathrm{MLP}(\hat{\mathbf{F}}_i^C), \mathbf{y}_i].
\end{equation}
For the third term, we aim to maximize the correlation between the label $\mathbf{y}$ and the emotional state features $\hat{\mathbf{F}}^C$, while minimizing the correlation between the label $\mathbf{y}$ and the interference features $\hat{\mathbf{F}}^I$. This is computed using the mutual information criterion:
\begin{equation}\label{equ:L_I}
\mathcal{L}_I = \frac{1}{n} \sum\nolimits_i I(\mathbf{y}_i; \hat{\mathbf{F}}_i^I | \hat{\mathbf{F}}_i^C).
\end{equation}
Thus, the final loss function in MMQ-Net is:
\begin{equation}\label{equ:L}
\mathcal{L} = \lambda_1 \mathcal{L}_R + \lambda_2 \mathcal{L}_C + \lambda_3 \mathcal{L}_I,
\end{equation}
where $\lambda_1$, $\lambda_2$, and $\lambda_3$ are non-negative trade-off parameters.

\section{Experiments and Results}

\subsection{Datasets}

We performed experiments utilizing two multi-modal physiological datasets, i.e., DEAP dataset~\cite{koelstra2011deap} and MAHNOB-HCI dataset~\cite{soleymani2011multimodal}. Both datasets elicit emotional responses through multimedia content.

\subsubsection{DEAP dataset} encompasses recordings from 32 individuals who were exposed to multimedia stimuli. Each subject participated in 40 sessions, during which they watched a one-minute music video per session. The recordings captured their physiological responses both before and during the viewing, with each session's data consisting of a 3-second pre-session phase and a 60-second session phase. This dataset includes 32 channels of EEG data alongside 8 channels of peripheral physiological signals, such as EOG, EMG, GSR, respiration, and temperature readings.

\subsubsection{MAHNOB-HCI dataset} gathers data from 27 participants subjected to multimedia stimuli. For this collection, each participant viewed 20 video clips while their physiological reactions were recorded across 20 trials. These clips varied in length from 34.9 seconds to 117 seconds, averaging 81.4 seconds with a standard deviation of 22.5 seconds. The dataset features 32 channels of EEG signals and 6 channels of peripheral physiological signals, including ECG, GSR, respiration, and temperature measurements.

\subsection{Experimental Setup}

The multi-masked querying transformer consists of an embedding layer, positional encoding, transformer encoder layers, and a classification head. Specifically, input features are projected into a 16-dimensional space using linear transformations, followed by adding positional encodings to maintain sequence order. In the multi-head attention mechanism, the head is set to 4 and the feed-forward dimension is 128. All experiments were implemented in PyTorch, utilizing an Adam optimizer with a learning rate of 6e-4, $\beta_1$ = 0.9, $\beta_2$ = 0.999, over 5000 epochs and a batch size of 1024. Hyper-parameters were selected based on cross-validation results, with $\lambda_1$ and $\lambda_2$ both set to 1, and $\lambda_3$ tuned to 0.01.

To comprehensively evaluate the performance of the proposed method, 
we designed an extensive experimental setup. Specifically, we assessed the effectiveness of CCA~\cite{hotelling1992relations_CCA}, KCCA~\cite{lopez2014randomized_KCCA}, DCCA~\cite{andrew2013deep_DCCA}, AE~\cite{lee2019audio_AE}, SMIL~\cite{ma2021smil}, ShaSpe~\cite{wang2023multi_ShaSpe}, TAE~\cite{cheng2024novel_TAE}, and our MMQ-Net on two categories, namely Valence and Arousal. The experiments were conducted with varying missing rates ranging from 0.1 to 0.7 to simulate real-world data conditions where missing values are common. This design allowed us to systematically compare the robustness and accuracy of each method under different levels of data incompleteness, providing insights into their suitability for handling missing data in emotional signal processing tasks.

\subsection{Comparison With the State-of-the-Art Methods}

Table~\ref{tab_DEAP} compares the performance of various methods at different missing rates on the DEAP dataset. As the missing rate increases, MMQ-Net's advantage becomes more evident. For example, at a 0.7 missing rate, MMQ-Net improves accuracy by up to 5.95\% for Valence and 4.44\% for Arousal over the next best method. This demonstrates MMQ-Net's robustness in handling high levels of data incompleteness, making it a strong candidate for emotion recognition in challenging conditions.

\begin{table}[tb]
\centering
\caption{Comparison under various missing rates on the DEAP dataset.}\label{tab_DEAP}
\begin{tabular}{ll cccccc}
\hline
 & & \multicolumn{5}{c}{\textbf{Missing rate}} \\
\cline{3-7}
\textbf{Category} & \textbf{Method} &  \textbf{0.0} & \textbf{0.1} & \textbf{0.3} & \textbf{0.5} & \textbf{0.7} \\
\hline
 & CCA & 66.55\% & 63.98\% & 61.91\% & 60.85\% & 59.43\% \\
 & KCCA & 67.20\% & 65.44\% & 64.68\% & 62.87\% & 61.05\% \\
 & DCCA & 71.64\% & 70.94\% & 70.79\% & 70.64\% & 68.77\% \\
 & AE & 75.88\% & 74.67\% & 74.37\% & 73.97\% & 71.85\% \\
Valence & SMIL & 88.14\% & 87.44\% & 83.75\% & 81.33\% & 76.99\% \\
 & ShaSpe & 88.04\% & 87.08\% & 84.71\% & 83.55\% & 81.94\% \\
 & TAE & 96.77\% & 92.38\% & 89.51\% & 87.24\% & 84.76\% \\
 & \textbf{MMQ-Net} & $\textbf{99.85\%}$ & $\textbf{95.71\%}$ & $\textbf{92.28\%}$ & $\textbf{90.97\%}$ & $\textbf{90.72\%}$ \\
 &  & ($\textbf{3.08\%}\uparrow$) & ($\textbf{3.33\%}\uparrow$) & ($\textbf{2.77\%}\uparrow$) & ($\textbf{3.73\%}\uparrow$) & ($\textbf{5.95\%}\uparrow$) \\
\cline{1-7}
 & CCA & 67.05\% & 64.28\% & 64.08\% & 62.31\% & 61.00\% \\
 & KCCA & 67.61\% & 66.90\% & 66.04\% & 64.73\% & 64.68\% \\
 & DCCA & 71.75\% & 69.48\% & 67.41\% & 67.81\% & 67.76\% \\
 & AE & 76.49\% & 75.53\% & 74.77\% & 72.86\% & 72.05\% \\
Arousal & SMIL & 88.35\% & 87.39\% & 83.96\% & 81.13\% & 79.21\% \\
 & ShaSpe & 86.98\% & 85.47\% & 84.16\% & 84.41\% & 81.74\% \\
 & TAE & 95.96\% & 92.23\% & 89.10\% & 87.18\% & 85.62\% \\
 & \textbf{MMQ-Net} & $\textbf{99.70\%}$ & $\textbf{95.86\%}$ & $\textbf{91.47\%}$ & $\textbf{90.62\%}$ & $\textbf{90.06\%}$ \\
 &  & ($\textbf{3.73\%}\uparrow$) & ($\textbf{3.63\%}\uparrow$) & ($\textbf{2.37\%}\uparrow$) & ($\textbf{3.43\%}\uparrow$) & ($\textbf{4.44\%}\uparrow$) \\
\hline
\end{tabular}
\end{table}

Table~\ref{tab_HCI} presents the performance comparison on the MAHNOB-HCI dataset. The results indicate that MMQ-Net consistently outperforms other methods across all missing rates. Specifically, MMQ-Net achieves the highest accuracy for both Valence and Arousal categories, with improvements ranging from 4.20\% to 6.76\% over the next best-performing method at different missing rates. This also demonstrates the robustness and effectiveness of MMQ-Net in handling varying levels of missing data.

\begin{table}[tb]
\centering
\caption{Comparison under various missing rates on the MAHNOB-HCI dataset.}\label{tab_HCI}
\begin{tabular}{ll cccccc}
\hline
 & & \multicolumn{5}{c}{\textbf{Missing rate}} \\
\cline{3-7}
\textbf{Category} & \textbf{Method} &  \textbf{0.0} & \textbf{0.1} & \textbf{0.3} & \textbf{0.5} & \textbf{0.7} \\
\hline
 & CCA & 69.46\% & 67.68\% & 66.90\% & 66.74\% & 65.19\% \\
 & KCCA & 70.86\% & 68.45\% & 67.99\% & 65.42\% & 64.88\% \\
 & DCCA & 73.50\% & 71.25\% & 68.30\% & 64.41\% & 61.46\% \\
 & AE & 77.78\% & 78.24\% & 75.68\% & 74.59\% & 72.96\% \\
Valence & SMIL & 87.80\% & 89.59\% & 85.47\% & 82.75\% & 80.89\% \\
 & ShaSpe & 86.01\% & 85.16\% & 84.07\% & 79.80\% & 78.32\% \\
 & TAE & 94.09\% & 90.91\% & 88.42\% & 86.64\% & 84.46\% \\
 & \textbf{MMQ-Net} & $\textbf{99.69\%}$ & $\textbf{96.04\%}$ & $\textbf{92.62\%}$ & $\textbf{91.69\%}$ & $\textbf{90.37\%}$ \\
 &  & ($\textbf{5.59\%}\uparrow$) & ($\textbf{5.13\%}\uparrow$) & ($\textbf{4.20\%}\uparrow$) & ($\textbf{5.05\%}\uparrow$) & ($\textbf{5.91\%}\uparrow$) \\
\cline{1-7}
 & CCA & 69.46\% & 65.66\% & 63.09\% & 59.36\% & 58.82\% \\
 & KCCA & 71.41\% & 68.61\% & 64.65\% & 63.87\% & 62.55\% \\
 & DCCA & 72.73\% & 72.57\% & 69.85\% & 68.61\% & 68.38\% \\
 & AE & 76.53\% & 72.42\% & 69.15\% & 65.58\% & 65.81\% \\
Arousal & SMIL & 90.29\% & 87.18\% & 83.53\% & 79.72\% & 77.23\% \\
 & ShaSpe & 86.17\% & 83.53\% & 84.38\% & 80.26\% & 79.10\% \\
 & TAE & 94.56\% & 90.75\% & 87.80\% & 86.95\% & 84.62\% \\
 & \textbf{MMQ-Net} & $\textbf{99.61\%}$ & $\textbf{95.80\%}$ & $\textbf{93.55\%}$ & $\textbf{92.46\%}$ & $\textbf{91.38\%}$ \\
 &  & ($\textbf{5.05\%}\uparrow$) & ($\textbf{5.05\%}\uparrow$) & ($\textbf{5.75\%}\uparrow$) & ($\textbf{5.52\%}\uparrow$) & ($\textbf{6.76\%}\uparrow$) \\
\hline
\end{tabular}
\end{table}

\subsection{Ablation Study}

To examine the effectiveness of each component in MMQ-Net, we carried out ablation studies by removing each of them from the whole framework under a missing rate of 0.3. As shown in Table~\ref{tab_Abs}, removing $\mathcal{L}_R$ leads to a noticeable drop in performance, with accuracy decreasing by approximately 5\% for Valence and by around 4\% for Arousal on both datasets. This highlights the critical importance of $\mathcal{L}_R$ for incomplete multi-modal learning. Similarly, when $\mathcal{L}_I$ is removed, the performance also decreases. These findings underscore the necessity of $\mathcal{L}_I$ in mitigating interference, thereby improving overall model performance.

\begin{table}[tb]
\centering
\caption{Ablation results of MMQ-Net.}\label{tab_Abs}
\begin{tabular}{c cccc}
\hline
 & DEAP & DEAP & HCI & HCI \\
 & (Valence) & (Arousal) & (Valence) & (Arousal) \\
\hline
MMQ-Net (w/o $\mathcal{L}_R$) & 86.43\% & 87.39\% & 87.72\% & 89.43\% \\
MMQ-Net (w/o $\mathcal{L}_I$) & 89.76\% & 89.05\% & 92.00\% & 92.15\% \\
MMQ-Net & 92.28\% & 91.47\% & 92.62\% & 93.55\% \\
\hline
\end{tabular}
\end{table}

\section{Conclusion}

We introduced the Multi-Masked Querying Network (MMQ-Net), a novel method for robust emotion recognition from incomplete multi-modal physiological signals. MMQ-Net combines multiple querying mechanisms to address challenges of missing data and interference. Modality queries reconstruct missing data, while category and interference queries distinguish relevant emotional features from noise. Extensive experiments on benchmark datasets show MMQ-Net's superior performance, particularly under high levels of missing data. Our results demonstrate its effectiveness in enhancing emotion recognition accuracy and reliability, making it a promising solution for emotion analysis and mental health monitoring in real-world applications.

\begin{credits}
\subsubsection{\ackname} This work is supported in part by the National Natural Science Foundation of China under Grants No.~62403449 and U2241210, and in part by the Shenzhen Science and Technology Program (CJGJZD20220517142000002), and in part by the National Key R\&D Program of China (2024YFC3607100).

\subsubsection{\discintname}
The authors have no competing interests to declare that are relevant to the content of this article.
\end{credits}
%
%
%
\bibliographystyle{splncs04}
\bibliography{Paper-0029}

\begin{thebibliography}{10}
\providecommand{\url}[1]{\texttt{#1}}
\providecommand{\urlprefix}{URL }
\providecommand{\doi}[1]{https://doi.org/#1}

\bibitem{andrew2013deep_DCCA}
Andrew, G., Arora, R., Bilmes, J., Livescu, K.: Deep canonical correlation
  analysis. In: International conference on machine learning. pp. 1247--1255.
  PMLR (2013)

\bibitem{cheng2024novel_TAE}
Cheng, C., Liu, W., Fan, Z., Feng, L., Jia, Z.: A novel transformer autoencoder
  for multi-modal emotion recognition with incomplete data. Neural Networks
  \textbf{172},  106111 (2024)

\bibitem{hotelling1992relations_CCA}
Hotelling, H.: Relations between two sets of variates. In: Breakthroughs in
  statistics: methodology and distribution, pp. 162--190. Springer (1992)

\bibitem{jia2024multi}
Jia, Z., Zhao, F., Guo, Y., Chen, H., Jiang, T., Center, B.: Multi-level
  disentangling network for cross-subject emotion recognition based on
  multimodal physiological signals. In: Proceedings of the Thirty-Third
  International Joint Conference on Artificial Intelligence. pp. 3069--3077
  (2024)

\bibitem{koelstra2011deap}
Koelstra, S., Muhl, C., Soleymani, M., Lee, J.S., Yazdani, A., Ebrahimi, T.,
  Pun, T., Nijholt, A., Patras, I.: Deap: A database for emotion analysis;
  using physiological signals. IEEE transactions on affective computing
  \textbf{3}(1),  18--31 (2011)

\bibitem{lee2019audio_AE}
Lee, H.C., Lin, C.Y., Hsu, P.C., Hsu, W.H.: Audio feature generation for
  missing modality problem in video action recognition. In: ICASSP 2019-2019
  IEEE International Conference on Acoustics, Speech and Signal Processing
  (ICASSP). pp. 3956--3960. IEEE (2019)

\bibitem{lian2023gcnet}
Lian, Z., Chen, L., Sun, L., Liu, B., Tao, J.: Gcnet: Graph completion network
  for incomplete multimodal learning in conversation. IEEE Transactions on
  pattern analysis and machine intelligence  \textbf{45}(7),  8419--8432 (2023)

\bibitem{lin2025brain}
Lin, Y., Xu, G.X., Liang, H., Wang, Y., Wan, F., Li, Y.: Brain region knowledge
  based dual-stream transformer for eeg emotion recognition. IEEE Transactions
  on Consumer Electronics  (2025)

\bibitem{liu2024contrastive}
Liu, R., Zuo, H., Lian, Z., Schuller, B.W., Li, H.: Contrastive learning based
  modality-invariant feature acquisition for robust multimodal emotion
  recognition with missing modalities. IEEE Transactions on Affective Computing
   (2024)

\bibitem{liu2023emotionkd}
Liu, Y., Jia, Z., Wang, H.: Emotionkd: a cross-modal knowledge distillation
  framework for emotion recognition based on physiological signals. In:
  Proceedings of the 31st ACM International Conference on Multimedia. pp.
  6122--6131 (2023)

\bibitem{lopez2014randomized_KCCA}
Lopez-Paz, D., Sra, S., Smola, A., Ghahramani, Z., Sch{\"o}lkopf, B.:
  Randomized nonlinear component analysis. In: International conference on
  machine learning. pp. 1359--1367. PMLR (2014)

\bibitem{luck2014introduction}
Luck, S.J.: An introduction to the event-related potential technique. MIT press
  (2014)

\bibitem{ma2021smil}
Ma, M., Ren, J., Zhao, L., Tulyakov, S., Wu, C., Peng, X.: {SMIL}: Multimodal
  learning with severely missing modality. In: Proceedings of the AAAI
  Conference on Artificial Intelligence. vol.~35, pp. 2302--2310 (2021)

\bibitem{miao2021automated}
Miao, M., Hu, W., Xu, B., Zhang, J., Rodrigues, J.J., De~Albuquerque, V.H.C.:
  Automated cca-mwf algorithm for unsupervised identification and removal of
  eog artifacts from eeg. IEEE Journal of Biomedical and Health Informatics
  \textbf{26}(8),  3607--3617 (2021)

\bibitem{miller2024quality}
Miller, C.R., McDonald, J.E., Grau, P.P., Wetterneck, C.T.: Quality of life in
  posttraumatic stress disorder: The role of posttraumatic anhedonia and
  depressive symptoms in a treatment-seeking community sample. Trauma Care
  \textbf{4}(1) (2024)

\bibitem{shou2024adversarial}
Shou, Y., Meng, T., Ai, W., Zhang, F., Yin, N., Li, K.: Adversarial alignment
  and graph fusion via information bottleneck for multimodal emotion
  recognition in conversations. Information Fusion  \textbf{112},  102590
  (2024)

\bibitem{shu2018review}
Shu, L., Xie, J., Yang, M., Li, Z., Li, Z., Liao, D., Xu, X., Yang, X.: A
  review of emotion recognition using physiological signals. Sensors
  \textbf{18}(7), ~2074 (2018)

\bibitem{soleymani2011multimodal}
Soleymani, M., Lichtenauer, J., Pun, T., Pantic, M.: A multimodal database for
  affect recognition and implicit tagging. IEEE transactions on affective
  computing  \textbf{3}(1),  42--55 (2011)

\bibitem{tang2024hierarchical}
Tang, J., Ma, Z., Gan, K., Zhang, J., Yin, Z.: Hierarchical multimodal-fusion
  of physiological signals for emotion recognition with scenario adaption and
  contrastive alignment. Information Fusion  \textbf{103},  102129 (2024)

\bibitem{vaswani2017attention}
Vaswani, A., Shazeer, N., Parmar, N., Uszkoreit, J., Jones, L., Gomez, A.N.,
  Kaiser, {\L}., Polosukhin, I.: Attention is all you need. Advances in neural
  information processing systems  \textbf{30} (2017)

\bibitem{wang2023multi_ShaSpe}
Wang, H., Chen, Y., Ma, C., Avery, J., Hull, L., Carneiro, G.: Multi-modal
  learning with missing modality via shared-specific feature modelling. In:
  Proceedings of the IEEE/CVF Conference on Computer Vision and Pattern
  Recognition. pp. 15878--15887 (2023)

\bibitem{wang2025generative}
Wang, S., Zhou, T., Shen, Y., Li, Y., Huang, G., Hu, Y.: Generative ai enables
  eeg super-resolution via spatio-temporal adaptive diffusion learning. IEEE
  Transactions on Consumer Electronics  (2025)

\bibitem{wilmer2021correlates}
Wilmer, M.T., Anderson, K., Reynolds, M.: Correlates of quality of life in
  anxiety disorders: review of recent research. Current psychiatry reports
  \textbf{23}, ~1--9 (2021)

\bibitem{zhang2024camel}
Zhang, L., Jin, L., Xu, G., Li, X., Xu, C., Wei, K., Liu, N., Liu, H.: Camel:
  capturing metaphorical alignment with context disentangling for multimodal
  emotion recognition. In: Proceedings of the AAAI Conference on Artificial
  Intelligence. vol.~38, pp. 9341--9349 (2024)

\bibitem{zhao2024feature}
Zhao, Y., Gu, J.: Feature fusion based on mutual-cross-attention mechanism for
  eeg emotion recognition. In: International Conference on Medical Image
  Computing and Computer-Assisted Intervention. pp. 276--285. Springer (2024)

\bibitem{zhu2023dynamic}
Zhu, Q., Zheng, C., Zhang, Z., Shao, W., Zhang, D.: Dynamic confidence-aware
  multi-modal emotion recognition. IEEE Transactions on Affective Computing
  \textbf{15}(3),  1358--1370 (2023)

\end{thebibliography}

\end{document}